\documentclass[letterpaper]{article} 
\usepackage{aaai2026}  
\usepackage{times}  
\usepackage{helvet}  
\usepackage{courier}  
\usepackage[hyphens]{url}  
\usepackage{graphicx} 
\usepackage{natbib}  
\usepackage{caption} 
\usepackage{multirow}
\usepackage{graphicx}
\usepackage{booktabs}
\usepackage{array}      
\usepackage{ragged2e}
\usepackage{tabularx}
\usepackage[inkscapelatex=false]{svg}
\usepackage[most]{tcolorbox}
\usepackage{subcaption}
\usepackage{makecell}

\usepackage{pifont}
\newcommand{\cmark}{\ding{51} }%
\newcommand{\xmark}{\ding{55} }%

\newcommand{\g}{GAICo}

\usepackage{algorithm}
\usepackage{algorithmic}

\usepackage{newfloat}
\usepackage{listings}
\DeclareCaptionStyle{ruled}{labelfont=normalfont,labelsep=colon,strut=off} 
\floatstyle{ruled}
\newfloat{listing}{tb}{lst}{}
\floatname{listing}{Listing}
\usepackage{xcolor}

\title{
    GAICo: A Deployed and Extensible Framework for \\ Evaluating Diverse and Multimodal Generative AI Outputs
}

\author{
    Nitin Gupta\textsuperscript{\rm 1}\thanks{Contact authors: niting@email.sc.edu \& biplav.s@sc.edu.},
    Pallav Koppisetti\textsuperscript{\rm 1},
    Kausik Lakkaraju\textsuperscript{\rm 1}, 
    Biplav Srivastava\textsuperscript{\rm 1}
}

\affiliations{
    \textsuperscript{\rm 1}University of South Carolina\\
    Columbia, South Carolina, USA\\
}

\begin{document}

\maketitle

\begin{abstract}
The rapid proliferation of Generative AI (GenAI) into diverse, high-stakes domains necessitates robust and reproducible evaluation methods. However, practitioners often resort to ad-hoc, non-standardized scripts, as common metrics are often unsuitable for specialized, structured outputs (e.g., automated plans, time-series) or holistic comparison across modalities (e.g., text, audio, and image). This fragmentation hinders comparability and slows AI system development.
To address this challenge, we present \g\ (\underline{G}enerative \underline{AI} \underline{Co}mparator): a deployed, open-source Python library that streamlines and standardizes GenAI output comparison. \g\ provides a unified, extensible framework supporting a comprehensive suite of reference-based metrics for unstructured text, specialized structured data formats, and multimedia (images, audio). Its architecture features a high-level API for rapid, end-to-end analysis, from multi-model comparison to visualization and reporting, alongside direct metric access for granular control.
We demonstrate \g's utility through a detailed case study evaluating and debugging complex, multi-modal AI Travel Assistant pipelines. 
\g\ empowers  AI researchers and developers to efficiently assess system performance, make evaluation reproducible, improve development velocity, and ultimately build more trustworthy AI systems, aligning with the goal of moving faster and safer in AI deployment. \textbf{Since its release on PyPI in Jun 2025, the tool has been downloaded over 16K times, across versions, by Dec 2025, demonstrating growing community interest.}
\end{abstract}

\begin{links}
    \link{Deployed Library}{pypi.org/project/GAICo}
    \link{Code}{github.com/ai4society/GenAIResultsComparator}
    \link{Documentation}{ai4society.github.io/projects/GenAIResultsComparatsor}
\end{links}

\section{Introduction}

The adoption of Generative AI (GenAI), particularly Large Language Models (LLMs), has been transformative across industries. However, this rapid integration has outpaced the development of standardized, accessible, and domain-appropriate evaluation tools. While standard Natural Language Processing (NLP) metrics like BLEU \cite{papineni-etal-2002-bleu} and ROUGE \cite{lin-2004-rouge} are useful, they are often insufficient for structured outputs like AI plans or time-series forecasts. In the absence of a unified tool, developers create bespoke evaluation scripts, leading to a fragmented and non-reproducible landscape that hinders the comparison and iterative improvement of AI systems.

\begin{figure}
    \centering
    \includegraphics[width=\linewidth]{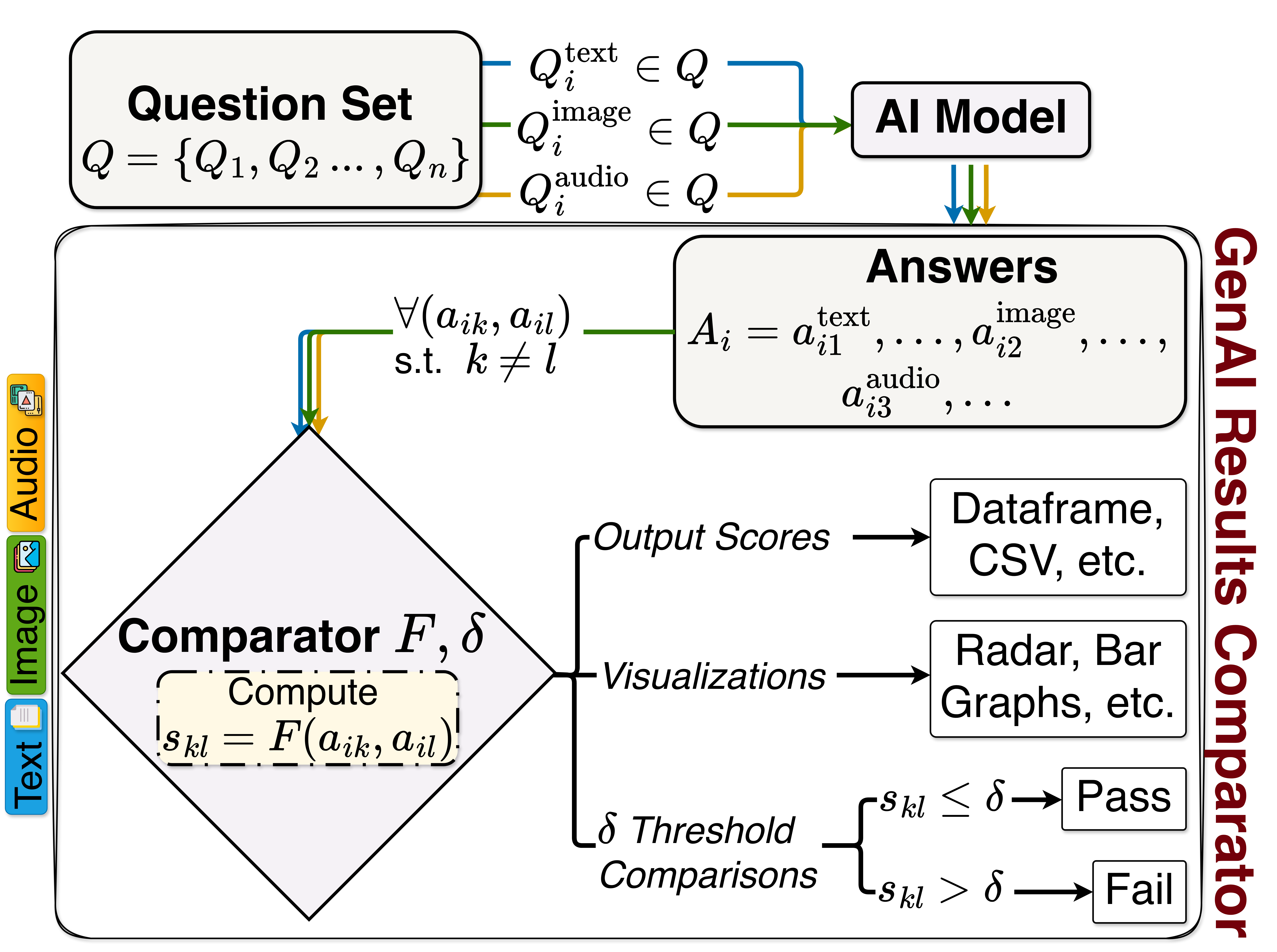}
    \caption{The multi-modal \g\ workflow. The framework processes answers from multi-modal (text, image, audio) AI models, computes pairwise similarity scores ($s_{kl}$), and constructs several outputs: raw data reports, visualizations, and pass/fail assessments against a threshold $\delta$. 
    (any distance function, or conversely, $1 - \text{similarity metric}$). 
    }
    \label{fig:simple-workflow}
\end{figure}

In response, we introduce \g\ (\underline{G}enerative \underline{AI} \underline{Co}mparator), a deployed and extensible open-source Python library built to address these challenges. \g\ provides a unifying framework that makes the application of diverse, reference-based metrics simple and reproducible across different data types (modalities). By providing a consistent interface, \g\ directly supports the iterative workflows common in applied AI development. According to a PyPI package statistics source \cite{pepy_gaico}, the library has been downloaded over 16,000 times in just over 6 months (Jun-Dec 2025), demonstrating its widespread and growing community interest. 

\begin{figure*}
    \centering
    \includegraphics[width=\linewidth]{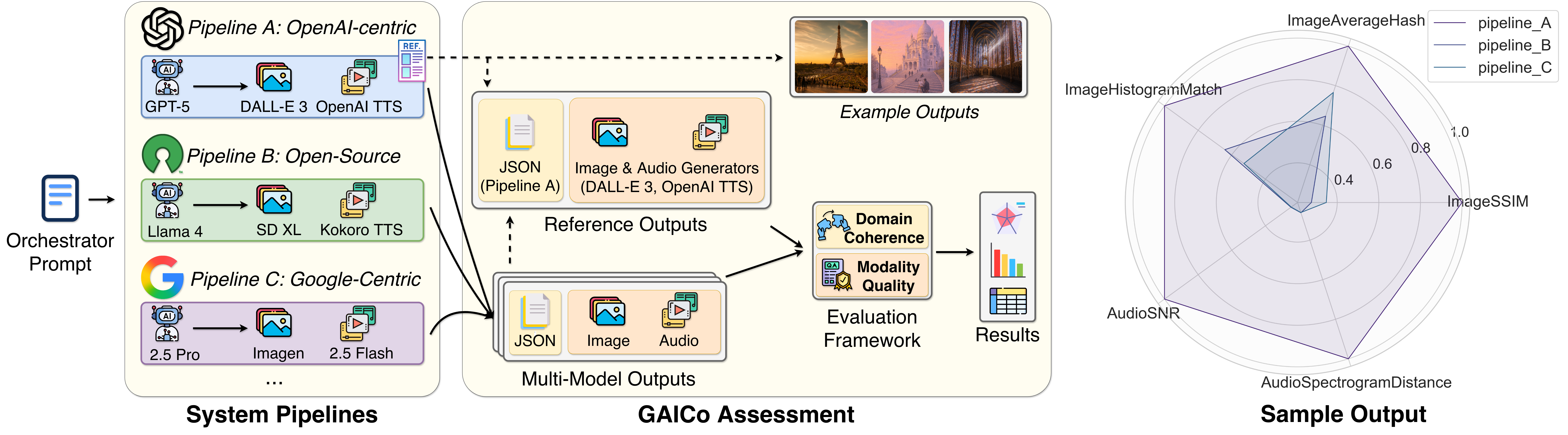}
    
    \caption{
        Illustration of \g\ for a composite AI system case study.
        (\textsl{Right}) A user uses LLMs to get results tailored to their needs and then analyzes them with \g. They start the process in three parallel pipelines that generate multi-modal outputs (JSON plan, image, audio).
        The user then performs a two-part evaluation: \textbf{(a) Plan Coherence}, comparing the JSON outputs of all pipelines against a single \textit{Baseline Plan Reference} (derived from Pipeline A's output) and \textbf{(b) Modality Quality}, comparing each pipeline's generated image and audio against \textit{per-pipeline references}. These per-pipeline references are generated by feeding the \textit{same prompts/scripts} from the respective pipeline's JSON output into baseline specialist generators (from Pipeline A).
        (\textsl{Left}) The output radar plot shows image and audio fidelity relative to references derived from each pipeline’s prompts.
        For more examples, see Table \ref{tab:example-notebooks}.
    }
    \label{fig:case_study_overview}
\end{figure*}

Common scenarios for usage of \g\ are evaluating an LLM's output along a range of metrics (Figure~\ref{fig:simple-workflow}) and output of multiple LLMs across a single metric (notebooks `example-1' and `example-2' in Table~\ref{tab:example-notebooks}). 
Consider the practical challenge
of building a 
significantly more complex
composite ``AI Travel Agent'' that uses an orchestrator LLM to generate an itinerary and specialist models to create corresponding images and audio summaries (System Pipelines in Figure \ref{fig:case_study_overview}). Evaluating different combinations of these components requires writing separate, ad-hoc scripts for each data modality. This process is slow, error-prone, and makes it difficult to pinpoint the root cause of a poor output: was it a flawed prompt from the orchestrator or poor execution by a specialist model? This exemplifies the need for a unified tool that can efficiently handle complex, multi-modal comparisons.

Our contributions, aligned with the ``Tools and Methodologies for Moving Faster and Safer'' track, are: 
(1)
        \noindent {\bf 
    A Unified and Extensible Framework:} A general framework for comparing GenAI outputs, with an object-oriented design that allows seamless integration of metrics for text, structured data (AI plans, time-series), and multimedia.
    (2)
    \noindent {\bf 
    A Streamlined Practitioner Workflow:} The \texttt{Experiment} class, which simplifies the end-to-end evaluation of complex AI systems by abstracting score calculation, visualization, and reporting into a few lines of code.
    (3)
    \noindent {\bf 
    Demonstrated Utility for Debugging Composite AI:} A case study on AI Travel Assistants (Figure \ref{fig:case_study_overview}) showcasing \g's effectiveness in performing nuanced, multi-stage evaluations to isolate performance issues.
    (4)
    \noindent {\bf 
    A Deployed and Maintained Tool:} A public tool on PyPI with comprehensive documentation, optional dependencies to manage installation size, and a robust testing suite to lower the barrier to adoption.

\section{Related Works}

\subsection{General-Purpose AI/NLP Evaluation Libraries}

General-purpose AI and Natural Language Processing (NLP) evaluation libraries provide foundational metrics for various tasks. For instance, the Hugging Face \texttt{evaluate} library \cite{wolf2019huggingface} offers a wide array of metrics for NLP and related tasks, providing a standardized \texttt{compute} method for individual metrics and integrating well within the Hugging Face ecosystem. Similarly, Scikit-learn \cite{scikit-learn} provides fundamental metrics for classification, regression, and clustering, adaptable for certain aspects of text evaluation (e.g., cosine similarity on TF-IDF vectors). Libraries like NLTK \cite{bird2009natural} and SpaCy \cite{Honnibal_spaCy_Industrial-strength_Natural_2020} offer foundational text processing capabilities and some basic metrics (e.g., BLEU in NLTK).

While indispensable, these libraries primarily offer individual text metrics and lack a unified framework for evaluating diverse, multi-modal GenAI outputs or streamlining multi-model comparisons. This fragmentation forces developers to manually integrate disparate tools. \g\ directly addresses this issue by providing a single, extensible API that unifies evaluation across diverse data types, significantly reducing development overhead and improving reproducibility.

A key differentiator for \g\ is its design choice to \textit{decouple evaluation from LLM inference}, positioning it as a post-hoc, reproducible comparison framework. This contrasts with general-purpose libraries like Hugging Face \texttt{evaluate} by offering a more integrated workflow and specialized metrics, and with end-to-end frameworks, e.g., Ragas \cite{es2024ragas}, DeepEval \cite{Ip_deepeval_2025}, that are tightly coupled with LLM APIs for ``LLM-as-a-judge'' evaluations. While these integrated frameworks are powerful, their coupling can introduce complexities like API costs, rate-limits, and nondeterminism. \g's focus on pre-generated outputs simplifies the evaluation pipeline, making it faster and more reliable for developers focused on comparative analysis. The \texttt{Experiment} class, discussed in Section 3\ref{sec:gaico_framework}, further streamlines this by offering a high-level abstraction for multi-model comparison, plotting, and reporting.
A detailed comparison of these approaches is provided in Section 1\ref{sec:appendix_comparison} of the Appendix.

\subsection{Domain-Specific Evaluation Approaches}

Evaluation in specialized AI domains often relies on custom tools, leading to fragmented practices. \g\ bridges this gap by integrating relevant metrics into a single, consistent framework.


    \paragraph{Automated Planning} 
    Evaluation in automated planning typically focuses on properties like plan length, cost, or state-space coverage, often requiring specialized planners or validators (e.g., PDDL parsers). However, when LLMs generate planning sequences, the output might not strictly adhere to formal planning languages, necessitating metrics that compare the \textit{content and order of actions} rather than strict plan validity. \g\ addresses this with \texttt{PlanningLCS} and \texttt{PlanningJaccard} metrics, which are designed to robustly compare action sequences, including concurrent actions, directly from LLM outputs. These are inspired by foundational work on measuring plan diversity and similarity \cite{srivastava2006finding,srivastava2007domain} and sequence analysis \cite{hirschberg1977algorithms}, adapted specifically for GenAI evaluation contexts.
    
    \paragraph{Time-Series Analysis} 
    Standard time-series evaluation metrics include Mean Absolute Error (MAE), Root Mean Squared Error (RMSE), or R-squared for forecasting accuracy; however, when comparing generated time-series \textit{shapes} or \textit{sequences of values} from LLMs, metrics like Dynamic Time Warping (DTW) \cite{berndt1994using} become crucial for handling variations in speed or phase. \g\ integrates \texttt{TimeSeriesDTW} and \texttt{TimeSeriesElementDiff} to provide robust, normalized comparisons of time-series data, regardless of their textual representation.
    
    \paragraph{Multimedia (Image/Audio)}
    Evaluation of generated images and audio often involves specialized libraries and metrics. For images, common metrics include Structural Similarity Index (SSIM) \cite{wang2004image}, Peak Signal-to-Noise Ratio (PSNR) \cite{sara2019image}, or perceptual hashes \cite{farid2021overview}. For audio, metrics like Signal-to-Noise Ratio (SNR) \cite{yuan2019signal}, Perceptual Evaluation of Speech Quality (PESQ) \cite{rix2001perceptual}, or Short-Time Objective Intelligibility (STOI) \cite{taal2010short} are used. While underlying libraries (e.g., scikit-image \cite{van2014scikit}, Pillow \cite{clark2015pillow}, librosa \cite{mcfee2015librosa}, scipy \cite{2020SciPy-NMeth}) provide these functionalities, \g\ unifies their application within its framework. The rise of multi-modal generative models, such as Meta's ImageBind \cite{girdhar2023imagebind}, further underscores the need for unified evaluation tools capable of assessing outputs across diverse modalities. This allows for consistent, normalized comparison of multimedia outputs alongside text and structured data, bridging the gap between general LLM evaluation and highly specialized domain-specific analysis.

\g's contribution lies in integrating these diverse, domain-specific metrics into a single, extensible library, accelerating development and improving the reliability of composite AI systems.


\section{The \texttt{\g}\ Framework}
\label{sec:gaico_framework}

The \g\ framework is architected to provide a robust, extensible, and user-friendly solution for evaluating GenAI outputs. Its design balances a high-level, streamlined workflow with the flexibility required for granular analysis, centered on three components: an extensible \texttt{BaseMetric} class, a comprehensive metric library, and a high-level \texttt{Experiment} class. 
A UML class diagram for \g\, can be viewed in Figure \ref{fig:gaico_uml} (Section 3 in the Appendix). 

\subsection{Core Concepts: BaseMetric and Extensibility}

At the heart of \g's architecture is the \texttt{BaseMetric} abstract class, which establishes a consistent foundation for all metrics. It mandates that every metric, regardless of data modality, implements a single core method: \texttt{calculate(generated\_texts, reference\_texts)}. This simple interface is key to the framework's flexibility, transparently handling various input formats (single items, lists, NumPy arrays) for efficient batch processing. This architecture makes \g\ highly extensible: a developer can introduce a new metric by simply inheriting from \texttt{BaseMetric} and implementing the \texttt{calculate()} method, immediately integrating it into the \g\ ecosystem.

\subsection{Comprehensive Metric Library}

\g\ provides a carefully designed library of metrics spanning textual analysis, structured data, and multimedia (audio and image). This comprehensive approach addresses the reality that modern GenAI applications produce complex, multi-modal outputs that traditional text-only tools cannot adequately assess. The library includes n-gram-based (BLEU, ROUGE), text similarity (Jaccard, Cosine), and semantic (BERTScore) metrics. For structured data, it offers specialized metrics for automated planning sequences (\texttt{PlanningLCS}) and time-series (\texttt{TimeSeriesDTW}). For multimedia, it integrates established metrics for image (SSIM, PSNR) and audio (SNR, Spectrogram Distance) quality.



\begin{figure*}[htbp]
\centering
\includegraphics[width=0.49\linewidth]{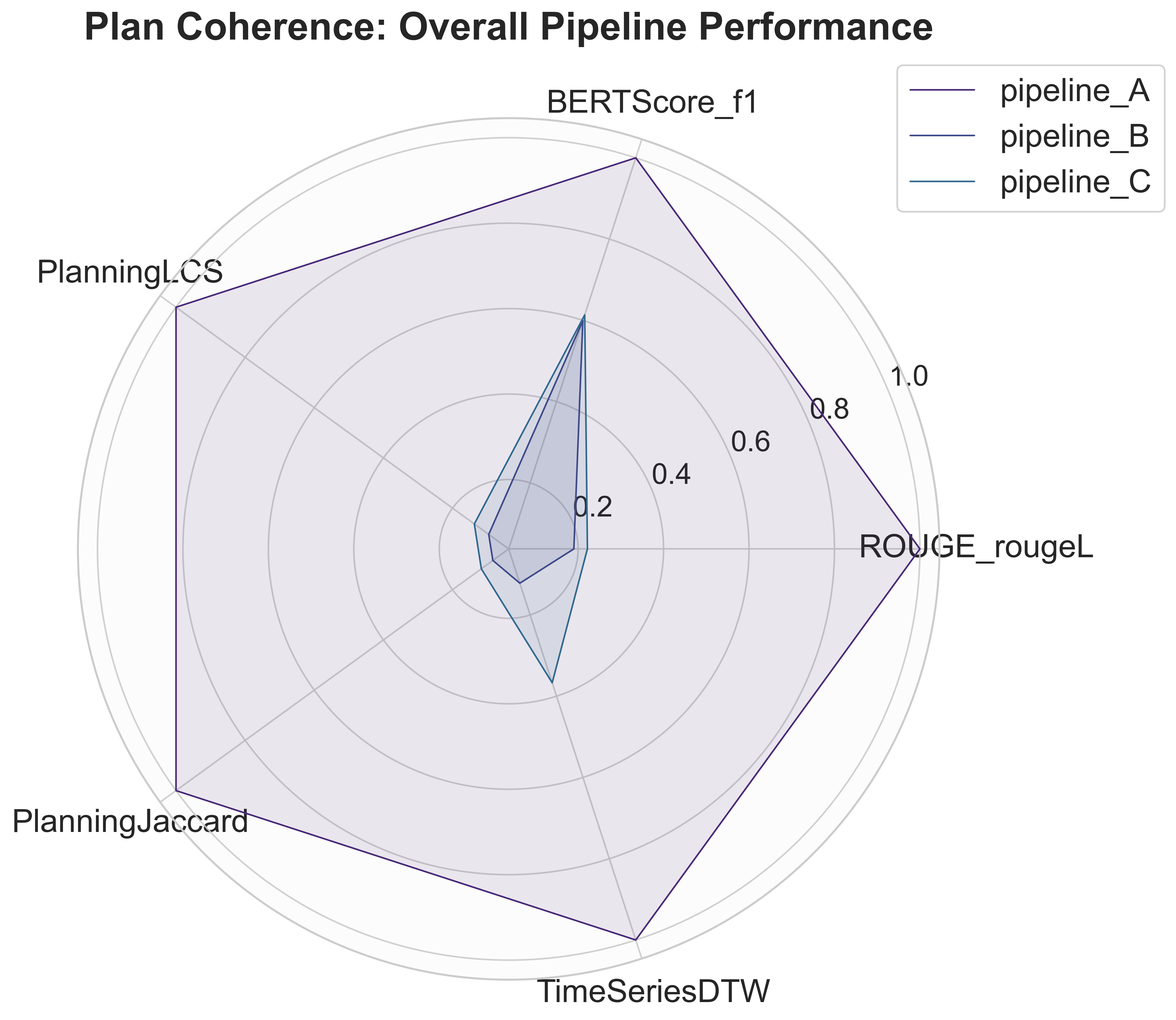} 
\includegraphics[width=0.49\linewidth]{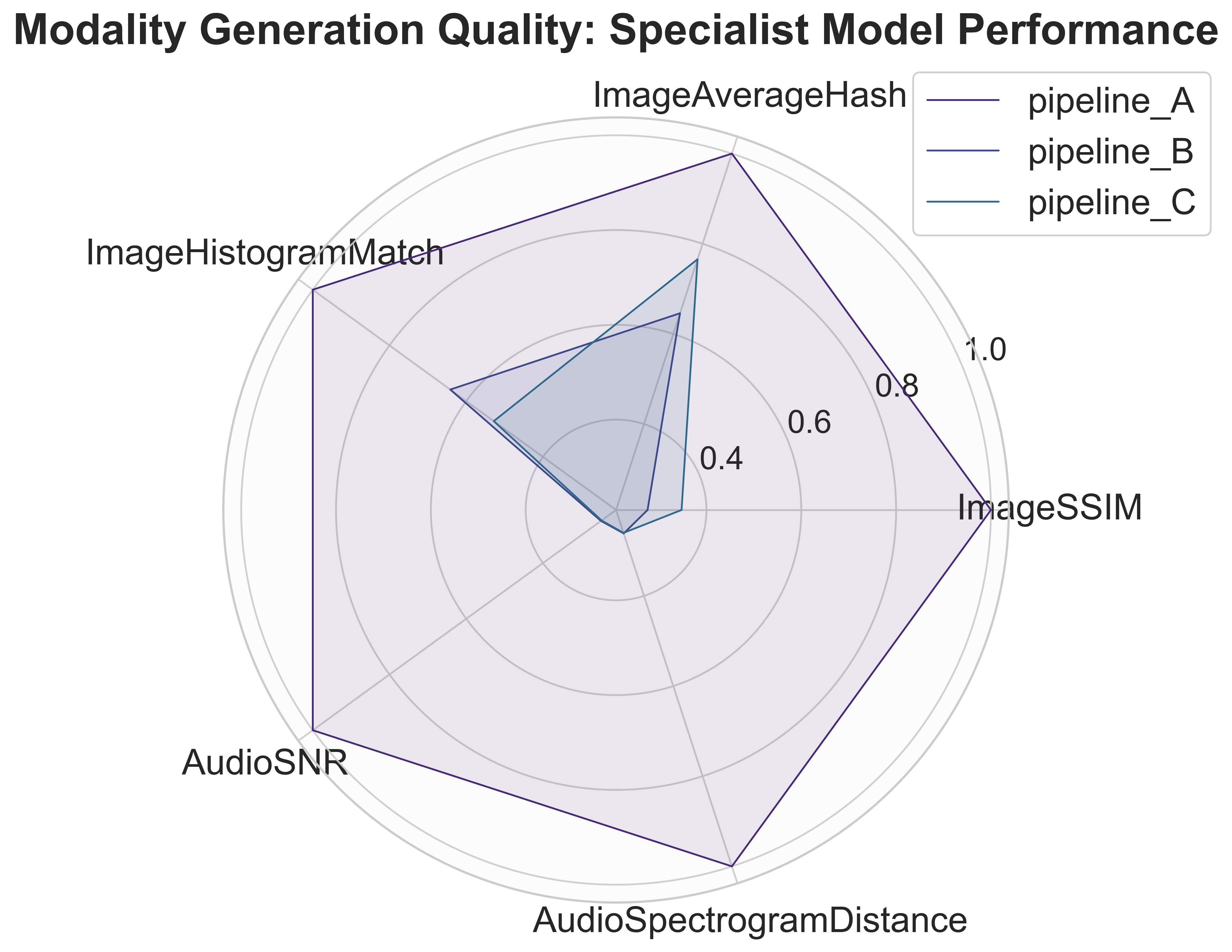}
\caption{
    Radar plots generated by \g\ comparing pipeline performance across various metrics.
    (\textsl{Right}) Modality Generation Quality, assessing the specialist models' fidelity against references generated from \textit{their own pipeline's prompts}. Each axis is a metric, and each line represents a pipeline's averaged score.
    (\textsl{Left}) Plan Coherence, showing the relative strengths of orchestrator LLMs against a universal human-curated reference.
}
\label{fig:radars}
\end{figure*}

\subsection{The Streamlined \texttt{Experiment} Class}

The \texttt{Experiment} class is a high-level API designed to streamline the entire evaluation workflow. It orchestrates the common task of comparing multiple model outputs against a single reference, encapsulating score calculation, visualization, and reporting into a few lines of code. Its \texttt{compare()} method automates the generation of scores, publication-ready plots (bar or radar), application of quality thresholds, and export of comprehensive CSV reports. By abstracting away boilerplate code, the \texttt{Experiment} class directly improves developer productivity and enables rapid, reproducible benchmarking.

\subsection{Deployment and Maintainability}

\g\ is a professionally engineered tool deployed on the Python Package Index (PyPI) for easy installation via \texttt{pip}. To balance functionality with accessibility, it uses optional dependencies (e.g., \texttt{pip install `gaico[bertscore]'}), minimizing the installation footprint. The project's quality is ensured through a comprehensive test suite (pytest \cite{pytest}), continuous integration (CI) for automated testing and deployment, and pre-commit hooks \cite{precommit_2025} for code consistency. Extensive documentation (MkDocs \cite{mkdocs_2025}), including runnable Jupyter notebooks and a full API reference, lowers the barrier to adoption for the applied AI community.

\section{Case Study: Evaluating Composite\\AI Travel Assistants}
\label{sec:case_study}

To demonstrate \g's utility in a realistic scenario, we evaluated three distinct, composite AI systems tasked with creating a multi-modal travel itinerary. This study highlights how \g\ enables a nuanced evaluation crucial for debugging and improving such systems. An overview of the case study is demonstrated in Figure \ref{fig:case_study_overview}. The code for the case study can be found in notebook number 7, `case\_study.ipynb', in Table~\ref{tab:example-notebooks}.

\begin{table*}[t]
\centering
\small
\setlength{\tabcolsep}{3.5pt}
\begin{tabular}{@{}lcccccccccc@{}}
\toprule
 & \multicolumn{5}{c}{\textbf{Plan Coherence Metrics}}
 & \multicolumn{5}{c}{\textbf{Modality Generation Metrics}} \\
\cmidrule(lr){2-6} \cmidrule(lr){7-11}
\textbf{Pipeline}
 & \textbf{$\text{ROUGE-L}_\textit{T}$}
 & \textbf{$\text{BERTScore-F1}_\textit{T}$}
 & \textbf{$\text{LCS}_\textit{Pl}$}
 & \textbf{$\text{Jaccard}_\textit{Pl}$}
 & \textbf{$\text{DTW}_{\textit{TS}}$}
 & \textbf{$\text{SSIM}_\textit{I}$}
 & \textbf{$\text{AvgHash}_\textit{I}$}
 & \textbf{$\text{HistMatch}_\textit{I}$}
 & \textbf{$\text{SNR}_\textit{A}$}
 & \textbf{$\text{SpecDist}_\textit{A}$} \\
\midrule
Pipeline~A
 & 1.000 & 1.000 & 1.000 & 1.000 & 1.000
 & 1.000 & 1.000 & 1.000 & 1.000 & 1.000 \\
Pipeline~B
 & 0.190 & 0.599 & 0.095 & 0.083 & 0.122
 & 0.276 & 0.646 & \textbf{0.642} & \textbf{0.249} & \textbf{0.261} \\
Pipeline~C
 & \textbf{0.222} & \textbf{0.613} & \textbf{0.137} & \textbf{0.117} & \textbf{0.367}
 & \textbf{0.347} & \textbf{0.766} & 0.528 & 0.247 & 0.260 \\
\bottomrule
\end{tabular}
\caption{Combined evaluation results. Plan coherence metrics (\textsl{left}) and modality generation metrics (\textsl{right}) are reported for each pipeline. Subscripts denote evaluation domains: \textit{T} = text, \textit{Pl} = planning, \textit{TS} = time series, \textit{I} = image, and \textit{A} = audio. Scores are averaged over three days; higher scores between Pipeline~B and Pipeline~C are highlighted.}
\label{tab:combined_results}
\end{table*}

\subsection{Setup and Methodology}

Our goal was to assess AI systems built from different combinations of state-of-the-art generative models. Each system, or ``pipeline'', consists of an orchestrator LLM that generates a travel plan in JSON format, which in turn provides prompts for downstream, specialized models for image and audio generation.

\paragraph{Master Orchestrator Prompt} All orchestrator LLMs received the same detailed prompt, designed to elicit a 3-day Paris itinerary in a structured JSON format, including fields for planning sequences, time-series, image prompts, and audio prompts (latter two for the specialist models). 
The full prompt is included in Section 2 of the Appendix.

We constructed three pipelines using different combinations of orchestrator LLMs and specialist image/audio models (Pipeline~A: OpenAI-centric, Pipeline~B: Open-Source, Pipeline~C: Google-centric). Each pipeline received an identical master prompt to generate a 3-day Paris itinerary in a structured JSON format, which included prompts for the downstream models.

A key challenge in composite systems is attributing failure. To address this, we designed a two-part evaluation strategy, seamlessly executed using \g:
\textbf{(1) Plan Coherence:} We evaluated the core travel planning ability of the orchestrator LLMs by comparing the generated JSON content (text, plans, budgets) from all pipelines against a single high-quality reference plan (from Pipeline~A).
\textbf{(2) Modality Generation Quality:} We evaluated the fidelity of the specialist models by comparing each pipeline's generated image and audio against a reference created using that pipeline's \textit{own} generated prompts. This isolates the specialist model's execution quality from the orchestrator's prompt-generation quality.



\subsection{Results and Analysis}
The quantitative results, summarized in Table~\ref{tab:combined_results} and visualized in Figure \ref{fig:radars}, demonstrate \g's value in accelerating AI development and improving system safety.

\paragraph{Orchestrator Performance (Plan Coherence)}
As shown in the left side of Table~\ref{tab:combined_results}, Pipeline~A's orchestrator (GPT-5), serving as the high-quality baseline, expectedly achieved perfect scores (1.000) across all plan coherence metrics. In contrast, both Pipeline~B (Llama 4) and Pipeline~C (Gemini 2.5 Pro) showed significantly lower performance, indicating substantial deviations from the baseline plan. As shown in the radar plot in Figure \ref{fig:radars} (left), Pipeline~C generally outperformed Pipeline~B across these metrics, suggesting a better ability to approximate the reference budget forecast.

\paragraph{Specialist Model Performance (Modality Generation Quality)}

The modality generation quality results (right of Table~\ref{tab:combined_results}) reveal the execution fidelity of the specialist models against their respective high-quality references. Pipeline~A's specialist models achieved perfect scores (1.000) as their outputs were used to define the modality-specific references. For image generation, Pipeline~C generally outperformed Pipeline~B across \texttt{Image SSIM} and \texttt{Image AverageHash}, indicating stronger structural and perceptual similarity to the baseline. However, Pipeline~B showed a slightly higher \texttt{Image HistogramMatch} score, suggesting its color distribution was closer to the baseline. This highlights the value of multi-metric evaluation for a nuanced understanding.

Focusing on the audio metrics, both Pipeline~B and Pipeline~C produced audio summaries with notably lower \texttt{Audio SNR} and \texttt{Audio SpectogramDist} compared to Pipeline~A. This indicates that while their orchestrators might provide valid scripts, their specialist audio models are not yet capable of matching the fidelity of the baseline OpenAI TTS. Figure \ref{fig:radars} (right) visually represents these performance differences.

\paragraph{Analysis}
This case study vividly demonstrates how \g\ facilitates ``moving faster and safer'' in the development of complex AI systems.

\textbf{Faster Development Cycles} Without \g, this multi-modal evaluation would require numerous, disparate scripts for JSON parsing, text analysis, plan validation, and multimedia comparison, followed by manual aggregation. \g\ streamlined this entire process into a single, unified workflow. This enables rapid iteration, allowing developers to swap components (e.g., an orchestrator LLM) and immediately receive quantitative feedback on the impact across all modalities, \textbf{drastically reducing the time and maintenance costs} associated with benchmarking.

\textbf{Improved Reliability and Safer Deployment} The two-part evaluation, enabled by \g, is critical for enhancing system \textbf{reliability}. By separating ``Plan Coherence'' from ``Modality Generation Quality'', we can precisely diagnose failures. For instance, the results show that Pipeline~B's poor performance stems from both a weak orchestrator (low scores in Table~\ref{tab:combined_results} left) and less-than-optimal specialist models (Table~\ref{tab:combined_results} right). This granular insight allows for targeted optimization, improving the orchestrator's planning logic or replacing a specific specialist model, which is essential for building trustworthy systems and preventing incidents caused by underperforming components. This directly contributes to the goal of producing \textbf{better AI solutions} by enabling a more rigorous and efficient development process.

\subsection{Demonstrated Use Cases and Documentation}

To maximize ease of use and adoption, key factors in a tool's real-world impact, \g\ is supported by extensive documentation and a comprehensive suite of example notebooks. These resources demonstrate the library's versatility across a range of applications, from simple quick-start guides to advanced, domain-specific evaluations in finance, planning, and multimedia.
Table \ref{tab:example-notebooks} summarizes the available examples, which are all publicly accessible in our code repository.

\begin{table*}[hbt!]
\centering
\setlength{\extrarowheight}{2pt}
\begin{tabularx}{\textwidth}{@{} p{0.1cm} l >{\RaggedRight}X @{}}
\toprule
& \textbf{Notebook Name} & \textbf{Description} \\
\midrule

\multicolumn{3}{@{}l}{\textbf{Quickstart Examples}} \\
01. & \emph{case\_study.ipynb} & Full code for the composite AI travel assistant evaluation presented in this paper. \\
02. & \emph{example-1.ipynb} & Evaluating multiple models using a single metric via the \texttt{calculate()} method. \\
03. & \emph{example-2.ipynb} & Evaluating a single model across multiple metrics using their \texttt{calculate()} methods. \\
04. & example-audio.ipynb & Evaluating AI-generated audio using specialized audio metrics. \\
05. & example-image.ipynb & Evaluating AI-generated images using specialized image metrics. \\
06. & example-structured\_data.ipynb & Comparing text-based metrics with specialized metrics for time-series and planning. \\
07. & quickstart.ipynb & A simple, quickstart workflow using \g's \texttt{Experiment} module. \\
\addlinespace
\midrule

\multicolumn{3}{@{}l}{\textbf{Intermediate Examples}} \\
08. & example-DeepSeek.ipynb & Evaluating DeepSeek for Point of View (POV) document analysis. \\
09. & example-audio\_data.ipynb & Advanced audio analysis for TTS and music generation tasks. \\
10. & example-election.ipynb & Evaluating models on sensitive election-related questions. \\
11. & example-finance.ipynb & Evaluating models on finance domain questions by iterating over a dataset. \\
12. & example-planning.ipynb & Evaluation of various travel plans using planning-specific metrics. \\
13. & example-recipes.ipynb & Evaluating models on recipe generation with parallelization via \textsc{joblib}. \\
14. & example-timeseries.ipynb & In-depth evaluation and perturbation of time-series data. \\

\midrule
\multicolumn{3}{@{}l}{\textbf{Advanced Examples}} \\
15. & example-llm\_faq.ipynb & Comparing various LLM responses (Phi, Mixtral, etc.) on a FAQ dataset. \\
16. & example-threshold.ipynb & Exploring default and custom thresholding techniques for LLM evaluation. \\
17. & example-viz.ipynb & Demonstrations of creating custom visualizations for evaluation results. \\
\bottomrule
\end{tabularx}
\setlength{\extrarowheight}{0pt}
\caption{
    \centering
    Summary of \g\ Example Notebooks. Those referenced in this paper are \emph{emphasized}. All available at:
    \newline
    https://github.com/ai4society/GenAIResultsComparator/tree/main/examples.
}
\label{tab:example-notebooks}
\end{table*}

\section{Conclusion}

In this section, we highlight the unique role of \g\ in evaluating LLM outputs in the context of prior work. Then, we discuss how it fits into the role of ``moving faster and safer'' in AI system development.

\subsection{\g\ and Its Novelty}
\g\ addresses a critical gap in GenAI evaluation by providing the first unified, extensible, and deployed framework for \textbf{reproducible} assessment across diverse data modalities. Recent studies have found that even with access to the original source code, many ML experiments cannot be reproduced, mainly due to missing details in documentation and inconsistent evaluation procedures \cite{lopresti2021reproducibility}. \g\ replaces this current landscape of ad-hoc, non-comparable scripts with a standardized approach, integrating metrics for text, structured data (planning, time-series), and multimedia. This standardization is \g's core novelty; it ensures that once a metric is chosen, its application and reporting are \textbf{consistent}, directly improving the \textbf{quality} and \textbf{productivity} of the evaluation process. As demonstrated in our case study, this unified approach enables nuanced debugging, such as distinguishing orchestrator faults from specialist model failures, which is crucial for enhancing the \textbf{reliability} of complex AI systems.

\subsection{\g\ as a Deployed Tool}
\g\ is engineered as a practical utility to help the community move \textbf{faster and safer}. Its design prioritizes \textbf{ease of use} and \textbf{adoption} through a simple PyPI installation, optional dependencies to manage installation size, comprehensive documentation, and a suite of runnable examples. This accessibility allows teams to \textbf{accelerate development} cycles; the streamlined \texttt{Experiment} class replaces disparate scripts with a single, consistent API, improving developer \textbf{productivity}. The framework enables a \textbf{safer deployment} of AI by facilitating precise failure analysis. By distinguishing a poor plan from a poor execution, \g\ improves system \textbf{reliability}, reduces long-term maintenance costs, and allows teams to mitigate weaknesses before they become incidents. Ultimately, \g\ is a deployed tool designed to have a significant \textbf{impact} by enabling the applied AI community to produce \textbf{better AI solutions} through rigorous, standardized, and efficient evaluation.

\subsection{Limitations and Future Work}
Despite its strengths, \g\ has some limitations in its current scope.
First, the current metric suite, while broad, primarily focuses on reference-based comparisons and does not yet include metrics for fairness, bias, toxicity, or operational aspects like latency and computational cost.
Furthermore, current visualizations are static plots (bar, radar). There is no built-in interactive dashboard or web-based UI for dynamic exploration of results.
Finally, while supporting structured data like planning and time-series, the library does not yet offer general-purpose comparison for arbitrary complex structured data (e.g., nested JSONs, knowledge graphs) or multi-turn conversational outputs.

We now discuss how one may extend the limitations in the future.
A key priority is to incorporate metrics for fairness, bias, toxicity, and operational statistics, providing developers with a more holistic toolkit for building safer and more responsible AI systems. \g's extensible design makes this easy.
Future work also includes exploring developing interactive dashboards or deeper integration with MLOps platforms (like MLflow \cite{Zaharia_Accelerating_the_Machine_2018}) to enable dynamic exploration and tracking of evaluation results.
Lastly, further work would also focus on adding general-purpose comparison capabilities for complex structured data (e.g., JSON diffing, graph similarity) and extending metrics to handle multi-turn conversational outputs.

\section*{Acknowledgments}

This work is partially supported by NSF Awards \#2454027, NAIRR250014, and Faculty Award by JP Morgan Research.

\bibliography{aaai2026}

\appendix
\section{Appendix}
\label{sec:appendix}

This appendix provides supplementary material to support the main paper. It includes a detailed comparison of our evaluation paradigm with existing tools, all the necessary artifacts for reproducing the AI Travel Assistant case study, and comprehensive descriptions of the diverse metrics implemented in our framework.

\begin{itemize}
    \item \textbf{Section 1\ref{sec:appendix_comparison}:} A comparative analysis of GAICo against other evaluation paradigms, highlighting its unique features.
    \item \textbf{Section 2\ref{sec:appendix_case_study}:} All artifacts for the AI Travel Assistant case study, including the orchestrator prompt and supplementary visualizations.
    \item \textbf{Section 3\ref{sec:appendix_metrics}:} Detailed descriptions of the textual, structured, and multimedia evaluation metrics available in GAICo.
\end{itemize}

\section{Comparison with Other Evaluation Paradigms}
\label{sec:appendix_comparison}

To better situate \g\ within the broader landscape of AI evaluation tools, Table \ref{tab:comparison} provides a detailed comparison against two dominant paradigms: general-purpose metric libraries, exemplified by Hugging Face \texttt{evaluate}, and end-to-end, LLM-integrated frameworks like Ragas and DeepEval. The table highlights \g's unique focus on post-hoc, reproducible evaluation of diverse data types (including specialized structured and perceptual multimedia metrics) and its streamlined workflow for multi-model comparative analysis, distinguishing it from libraries focused on standard NLP/CV tasks and frameworks tightly coupled with LLM inference.

\begin{table*}[t]
\centering
\caption{Comparison of GAICo with other AI Evaluation Paradigms}
\label{tab:comparison}
\resizebox{\linewidth}{!}{%
\begin{tabular}{@{}llll@{}}
\toprule
\textbf{Feature / Approach} & \textbf{GAICo} & \textbf{Hugging Face \texttt{evaluate}} & \textbf{LLM-Integrated Frameworks} \\
\midrule
\addlinespace[0.3em]
\textbf{Evaluation Paradigm} & Post-hoc, reference-based comparison & Post-hoc computation of individual metrics & End-to-end, often reference-free (LLM-as-a-judge) \\
& of pre-generated outputs & & \\
\addlinespace[0.3em]
\textbf{Primary Scope} & Text, Structured Data (Planning/TS), & General NLP \& CV task metrics & RAG-specific metrics (e.g., Faithfulness, Relevancy) \\
& Perceptual Multimedia & (e.g., BLEU, Accuracy, COCO) & \\
\addlinespace[0.3em]
\textbf{High-level Workflow} & \cmark (\texttt{Experiment} class for multi-model & \xmark (Focus on single metric \texttt{compute} calls) & \cmark (Orchestrates generation \& evaluation pipeline) \\
& analysis, plotting, reporting) & & \\
\addlinespace[0.3em]
\textbf{Decoupled from Generation} & \cmark (Ensures reproducibility, avoids & \cmark (Operates on pre-generated data) & \xmark (Tightly coupled with LLM inference APIs) \\
& API costs/latency during eval) & & \\
\addlinespace[0.3em]
\textbf{Built-in Visualization} & \cmark (Plots for comparative analysis) & \xmark (Requires external plotting libraries) & \xmark Varies (Some offer dashboards, not general plotting) \\
\addlinespace[0.3em]
\textbf{Extensibility} & \cmark (Inherit \texttt{BaseMetric}) & \cmark (Inherit \texttt{Metric}) & \cmark (Often supports custom metrics/prompts) \\
\bottomrule
\end{tabular}%
}
\end{table*}

\section{Case Study Artifacts}
\label{sec:appendix_case_study}

This section contains the material necessary for the \g\ AI Travel Assistant case study (Section 4\ref{sec:case_study}). The remaining scripts can be found in the GAICo repository.

\subsection{Orchestrator Prompt}

The orchestrator prompt shown in Figure~\ref{fig:travel-reaching-definitions} is the single system instruction fed to the top-level LLM that steers the entire AI Travel Assistant pipeline.  It fulfils three roles: (i) it elicits a structured \texttt{JSON} itinerary that can be parsed reliably by downstream code; (ii) it embeds sub-prompts (\textit{image\_prompt} and \textit{audio\_script}) that seed the specialist image and TTS generators, thereby keeping all modalities semantically aligned; and (iii) it enforces output-format contracts that make automated evaluation with \g\ possible.

\begin{figure}[htbp]
\centering
\scriptsize
\begin{tcolorbox}[
    colframe=black!0, 
    colback=gray!5, 
    coltitle=black,
    boxrule=0.5pt, 
    title=Orchestrator Prompt, 
    fonttitle=\bfseries,
    width=\columnwidth, 
    rounded corners, 
    colbacktitle=gray!20
]
You are a sophisticated AI Travel Assistant Orchestrator. Your task is to generate a travel plan based on the user's request. \\

Your output MUST be a single, valid JSON object and nothing else. Do not include any introductory text, explanations, or markdown formatting like \textasciigrave\textasciigrave\textasciigrave json. \\

The user's request is: ``A 3-day Paris trip for a first-time visitor on a moderate budget.'' \\

Please generate a JSON object with a root key ``trip\_plan'' which is an array of objects, one for each day. Each daily object must contain the following keys:

\begin{itemize}
    \item[--] ``day'': (Integer) The day number.
    \item[--] ``day\_plan\_text'': (String) A descriptive, paragraph-long summary of the day's activities.
    \item[--] ``day\_plan\_sequence'': (String) A comma-separated sequence of actions representing the plan, e.g., "visit(location), eat(restaurant), see(landmark)".
    \item[--] ``day\_budget\_euros'': (Number) An estimated budget in Euros for the day.
    \item[--] ``image\_prompt'': (String) A detailed, artistic, and descriptive prompt for an image generation model, capturing the essence of the day's main landmark.
    \item[--] ``audio\_script'': (String) A concise, engaging 15-second summary of the day's activities, written as a script for a text-to-speech model.
\end{itemize}

Now, generate the complete JSON object for the 3-day trip.

\end{tcolorbox}
\caption{Prompt used by the \textit{orchestrator LLM}.  The prompt enforces JSON output, supplies slot-level requirements for each day, and instructs downstream specialist generators (image and TTS) via embedded sub-prompts.}
\label{fig:travel-reaching-definitions}
\end{figure}

\subsection{Supplementary Bar Charts}
\begin{figure*}[t]
    \centering
    \includegraphics[width=\linewidth]{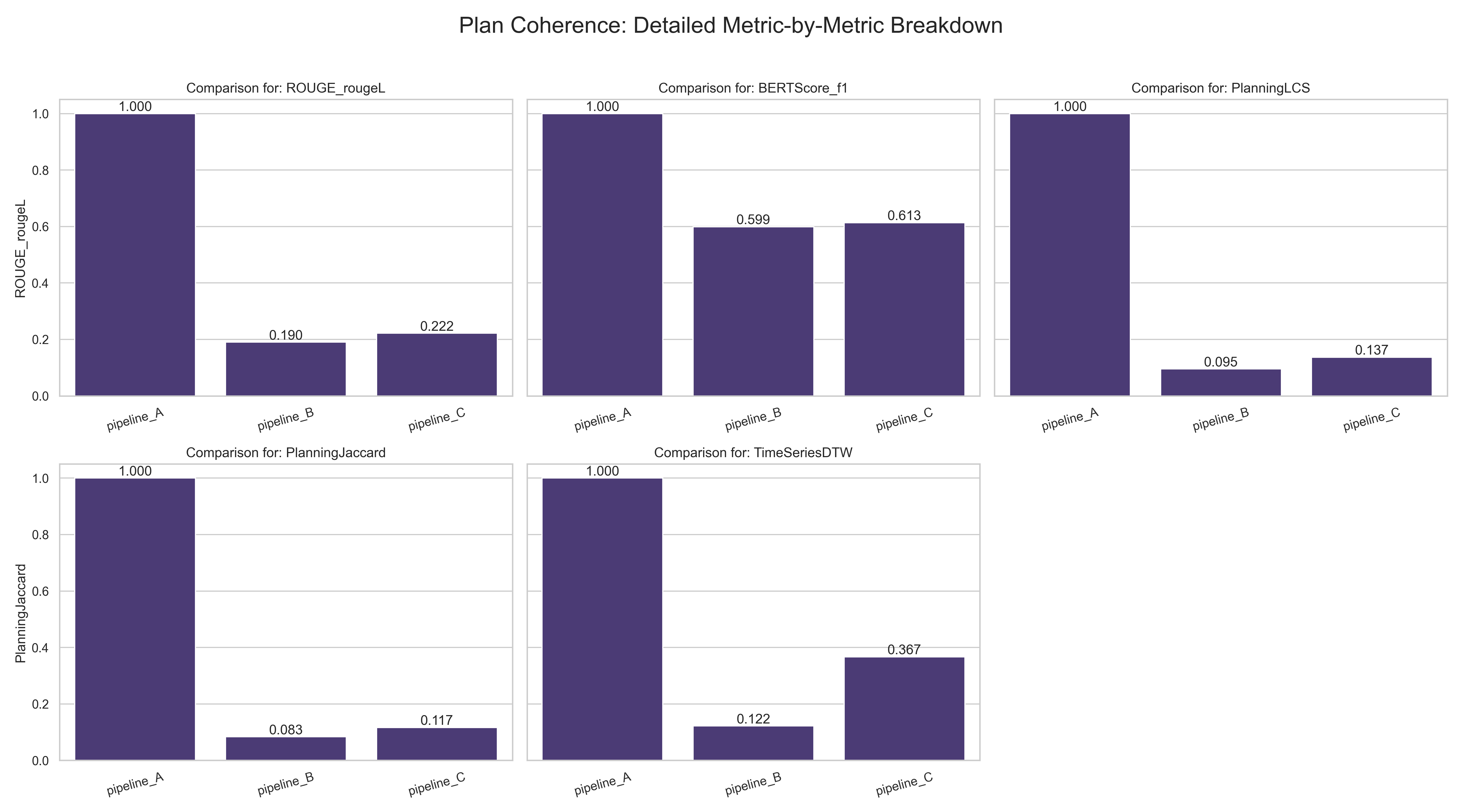}
    \caption{Per-metric comparison of \textbf{plan coherence} across the
    three pipelines. Higher is better.}
    \label{fig:plan_coherence_bars}
\end{figure*}

\begin{figure*}[t]
    \centering
    \includegraphics[width=\linewidth]{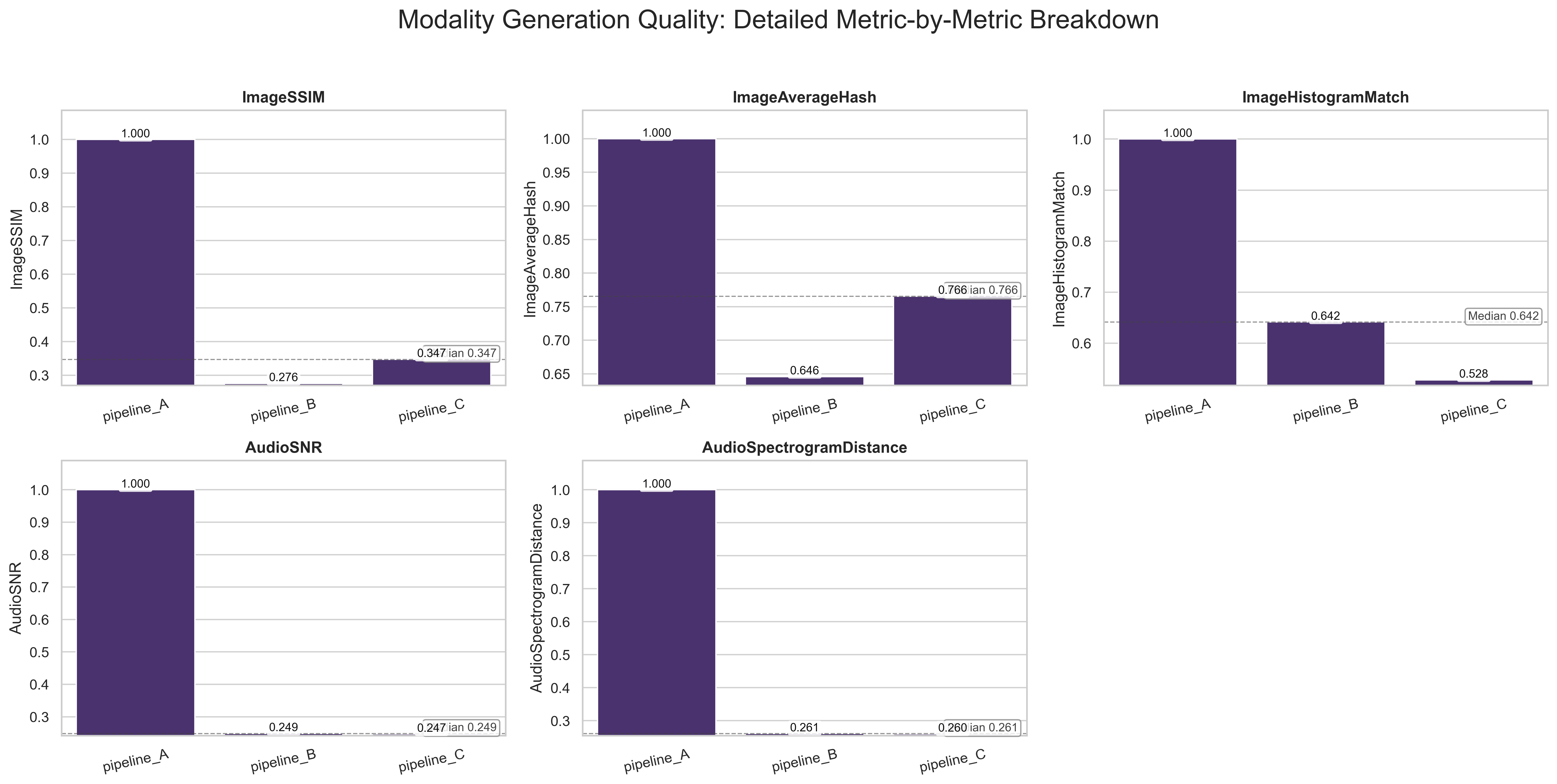}
    \caption{Per-metric comparison of \textbf{modality-generation
    quality} (images and audio) across the three pipelines.}
    \label{fig:modality_quality_bars}
\end{figure*}

To complement the radar plots shown in the main paper, Figures~\ref{fig:plan_coherence_bars} and \ref{fig:modality_quality_bars} provide a \emph{metric-by-metric} breakdown for the three pipelines evaluated in the AI Travel Assistant case study.

\paragraph{Figure~\ref{fig:plan_coherence_bars}: Plan-coherence view.}

The five bars in each subplot correspond to the specialised textual or structured metrics discussed. A score of 1.0 denotes perfect agreement with the \textit{baseline reference plan} produced by Pipeline~A.  The chart shows that Pipeline~C consistently outperforms Pipeline~B, hinting that the models in its pipeline improved temporal consistency.

\paragraph{Figure~\ref{fig:modality_quality_bars}: Modality-quality view.}

Here, we isolate the image and audio specialist generators. Because each pipeline’s own outputs serve as its reference, a perfect score of 1.0 appears only for Pipeline~A, which supplies those references. Lower bars for Pipelines~B and~C quantify \emph{perceptual} deviations. Pipeline~C’s higher \texttt{Image\_AverageHash} score versus Pipeline~B indicates better content preservation despite equal SSIM degradation. Both alternatives exhibit similar audio issues, as reflected by overlapping \texttt{Audio\_SNR} and \texttt{Audio\_SpectrogramDist} values.

Together, the two figures pinpoint that most performance regressions stem from plan-generation weaknesses rather than the downstream modality specialists, thereby guiding future remediation work.

\section{Detailed Metric Library Descriptions}
\label{sec:appendix_metrics}

\begin{figure*}
    \centering
    \includegraphics[width=1\linewidth]{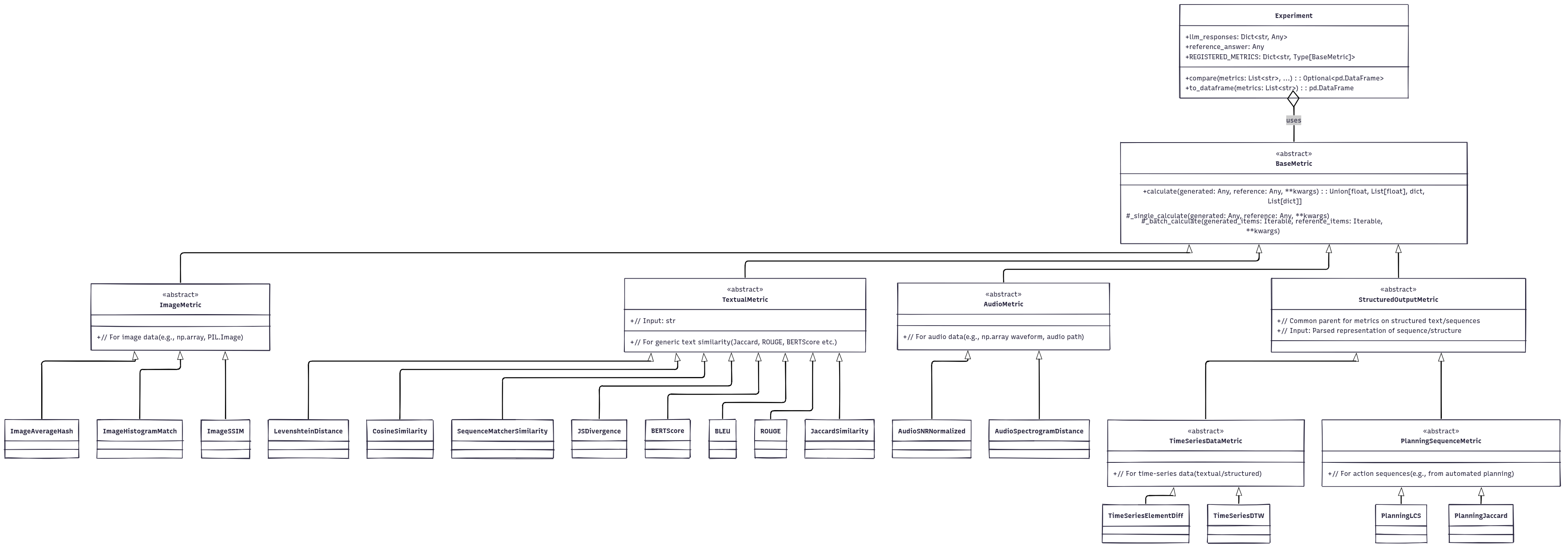}
    \caption{UML Class Diagram showcasing \g's capabilities and broad multi-media support.}
    \label{fig:gaico_uml}
\end{figure*}

This section provides detailed descriptions of the metrics available in \g, summarized in Table \ref{tab:metrics-comprehensive}. For an overview UML class diagram of \g, view Figure \ref{fig:gaico_uml}.

\subsection{Textual Evaluation Framework}
Our textual metrics encompass three complementary evaluation paradigms. 
\begin{itemize}
    \item \textbf{N-gram based metrics} include \texttt{BLEU} for translation-quality assessment, \texttt{ROUGE} variants (rouge1, rouge2, rougeL) for summarization, and \texttt{JSDivergence} Jensen-Shannon Divergence for vocabulary distribution analysis.
    \item \textbf{Text similarity metrics} provide orthogonal perspectives. \texttt{JaccardSimilarity} measures set-based word overlap, \texttt{CosineSimilarity} captures semantic relationships through TF-IDF vector space modeling, \texttt{LevenshteinDistance} quantifies edit-based differences, and \texttt{SequenceMatcherSimilarity} provides character-level analysis.
    \item \textbf{Semantic Similarity} is implemented via \texttt{BERTScore}, which leverages pre-trained BERT embeddings to compute contextual similarity, addressing the limitations of surface-level metrics.
\end{itemize}

\subsection{Structured Data Evaluation}
We developed specialized metrics for critical structured data applications.
\begin{itemize}
    \item \textbf{Planning sequence evaluation} addresses automated planning outputs. \texttt{PlanningLCS} implements longest common subsequence matching that respects temporal ordering and handles concurrent actions. \texttt{PlanningJaccard} focuses on action set completeness, ignoring ordering.
    \item \textbf{Time-series evaluation} tackles temporal data. \texttt{TimeSeriesElementDiff} implements a weighted comparison of keyed time-points. \texttt{TimeSeriesDTW} leverages Dynamic Time Warping to compare value sequences, effectively handling phase-shifted data.
\end{itemize}

\subsection{Multimedia Evaluation Capabilities}
\begin{itemize}
    \item \textbf{Audio evaluation} includes \texttt{AudioSNRNormalized}, which provides a normalized Signal-to-Noise Ratio on a [0, 1] scale, and \texttt{AudioSpectrogramDistance}, which evaluates similarity through frequency-time domain analysis using STFT.
    \item \textbf{Image evaluation} provides a suite of perceptual and structural metrics. \texttt{SSIM} measures perceived quality through structural information, luminance, and contrast. \texttt{PSNR} quantifies pixel-wise fidelity. \texttt{AverageHash} generates perceptual hashes for content similarity, and \texttt{HistogramMatch} evaluates color distribution similarity.
\end{itemize}

\begin{table*}[t]
\centering
\caption{\g's Comprehensive Metric Library}
\label{tab:metrics-comprehensive}
\newcolumntype{P}[1]{>{\RaggedRight\arraybackslash}p{#1}}
\setlength{\extrarowheight}{1pt}
\resizebox{\linewidth}{!}{
{\footnotesize
\begin{tabular}{ @{} l l l l @{} }
\toprule
\textbf{Category} & \textbf{Metric} & \textbf{Description} & \textbf{Primary Use Cases} \\
\midrule
\multirow{8}{*}{\textbf{Textual}}
& BLEU & N-gram precision with brevity penalty & Translation, text generation quality \\
& ROUGE & Recall-oriented n-gram and LCS matching & Summarization, content coverage \\
& JSDivergence & Symmetric vocabulary distribution comparison & Bias detection, style consistency \\
& JaccardSimilarity & Set-based word overlap analysis & Keyword extraction evaluation \\
& CosineSimilarity & TF-IDF vector space comparison & Document semantic similarity \\
& LevenshteinDistance & Character-level edit distance & Error analysis, text correction \\
& SequenceMatcher & Python difflib-based similarity & Code generation, precise matching \\
& BERTScore & BERT embedding-based semantic similarity & Deep semantic understanding \\
\midrule
\multirow{4}{*}{\textbf{Structured}}
& PlanningLCS & LCS with concurrent action support & Action sequence evaluation \\
& PlanningJaccard & Set-based action completeness & Plan coverage assessment \\
& TimeSeriesElementDiff & Weighted temporal data comparison & Forecasting evaluation \\
& TimeSeriesDTW & Dynamic time warping alignment & Phase-shifted sequence analysis \\
\midrule
\multirow{2}{*}{\textbf{Audio}}
& AudioSNRNormalized & Signal-to-noise ratio assessment & Audio quality evaluation \\
& AudioSpectrogramDistance & STFT-based frequency analysis & Audio generation evaluation \\
\midrule
\multirow{4}{*}{\textbf{Image}}
& SSIM & Structural similarity index & Image quality assessment \\
& PSNR & Peak signal-to-noise ratio & Image fidelity evaluation \\
& AverageHash & Perceptual hash for content similarity & Duplicate detection, content matching \\
& HistogramMatch & Color distribution similarity analysis & Global color profile comparison \\
\bottomrule
\end{tabular}
}
}
\setlength{\extrarowheight}{0pt}
\end{table*}

\end{document}